%% file: main.tex
\documentclass[conference]{IEEEtran}

\usepackage{cite}
\usepackage{amsmath,amssymb,amsfonts}
\usepackage{algorithmic}
\usepackage{graphicx}
\usepackage{textcomp}
\usepackage{xcolor}
\def\BibTeX{{\rm B\kern-.05em{\sc i\kern-.025em b}\kern-.08em
    T\kern-.1667em\lower.7ex\hbox{E}\kern-.125emX}}

\usepackage{multirow}
\usepackage{colortbl}
\definecolor{RowColor}{rgb}{0.93, 0.93, 1}
\usepackage{xcolor}
\usepackage{booktabs}
\usepackage{subcaption}

\begin{document}

\title{AdaSports-Traj: Role- and Domain-Aware Adaptation for Multi-Agent Trajectory Modeling in Sports}
\author{
\IEEEauthorblockN{Yi Xu, Yun Fu}
\IEEEauthorblockA{Northeastern University, Boston, USA\\
xu.yi@northeastern.edu, yunfu@ece.neu.edu}
}

\maketitle

\input{section/1_abstract}

\begin{IEEEkeywords}
Multi-agent trajectory modeling, sports analytics, hierarchical contrastive learning, disentangled representations.
\end{IEEEkeywords}

\input{section/2_introduction}
\input{section/3_relatedwork}
\input{section/4_method}
\input{section/5_experiments}
\input{section/6_conclusion}

\bibliographystyle{IEEEtran}
\bibliography{IEEEexample}

\end{document}

%% file: section/1_abstract.tex
\begin{abstract}
Trajectory prediction in multi-agent sports scenarios is inherently challenging due to the structural heterogeneity across agent roles (e.g., players vs. ball) and dynamic distribution gaps across different sports domains. Existing unified frameworks often fail to capture these structured distributional shifts, resulting in suboptimal generalization across roles and domains. We propose AdaSports-Traj, an adaptive trajectory modeling framework that explicitly addresses both intra-domain and inter-domain distribution discrepancies in sports. At its core, AdaSports-Traj incorporates a Role- and Domain-Aware Adapter to conditionally adjust latent representations based on agent identity and domain context. Additionally, we introduce a Hierarchical Contrastive Learning objective, which separately supervises role-sensitive and domain-aware representations to encourage disentangled latent structures without introducing optimization conflict. Experiments on three diverse sports datasets, Basketball-U, Football-U, and Soccer-U, demonstrate the effectiveness of our adaptive design, achieving strong performance in both unified and cross-domain trajectory prediction settings.
\end{abstract}

%% file: section/2_introduction.tex
\section{Introduction}
Understanding the movement patterns of multiple agents is a fundamental task in many real-world applications. Among various domains, sports scenes present structured yet highly dynamic multi-agent interactions, typically involving both players and a ball under explicit tactical and rule-based constraints. Accurately modeling agent trajectories in sports has become increasingly important for downstream applications such as strategy analysis~\cite{capellera2025unified}, simulation~\cite{li2025artificial}, augmented broadcast~\cite{han2007real}, and autonomous coaching systems~\cite{sha2018interactive,xu2024vlm}.

Despite notable advancements in trajectory modeling, much of the focus has centered on trajectory prediction, driven in part by the growing interest in autonomous driving. Numerous works~\cite{xu2020cf, xu2021tra2tra, xu2022remember, rowe2023fjmp, mao2023leapfrog, jiang2023motiondiffuser, chen2023unsupervised, zhou2023query, gu2023vip3d, aydemir2023adapt, bae2023eigentrajectory, shi2023trajectory, chen2023traj, seff2023motionlm, park2023leveraging, xu2023eqmotion, park2024improving, xu2024adapting} have addressed this task by predicting future trajectories from fixed-length, fully observed past sequences. 
However, such assumptions rarely hold in actual sports settings, where occlusions and tracking failures frequently lead to incomplete observation sequences with varying lengths across different scenes and tasks. 
To address these issues, recent methods~\cite{xu2023uncovering, qin2023multiple, xu2024adapting} adopt generative frameworks that introduce masking mechanisms to handle missing data, achieving promising results on benchmarks.

\begin{figure}[t]
\centering
\includegraphics[width=0.98\linewidth]{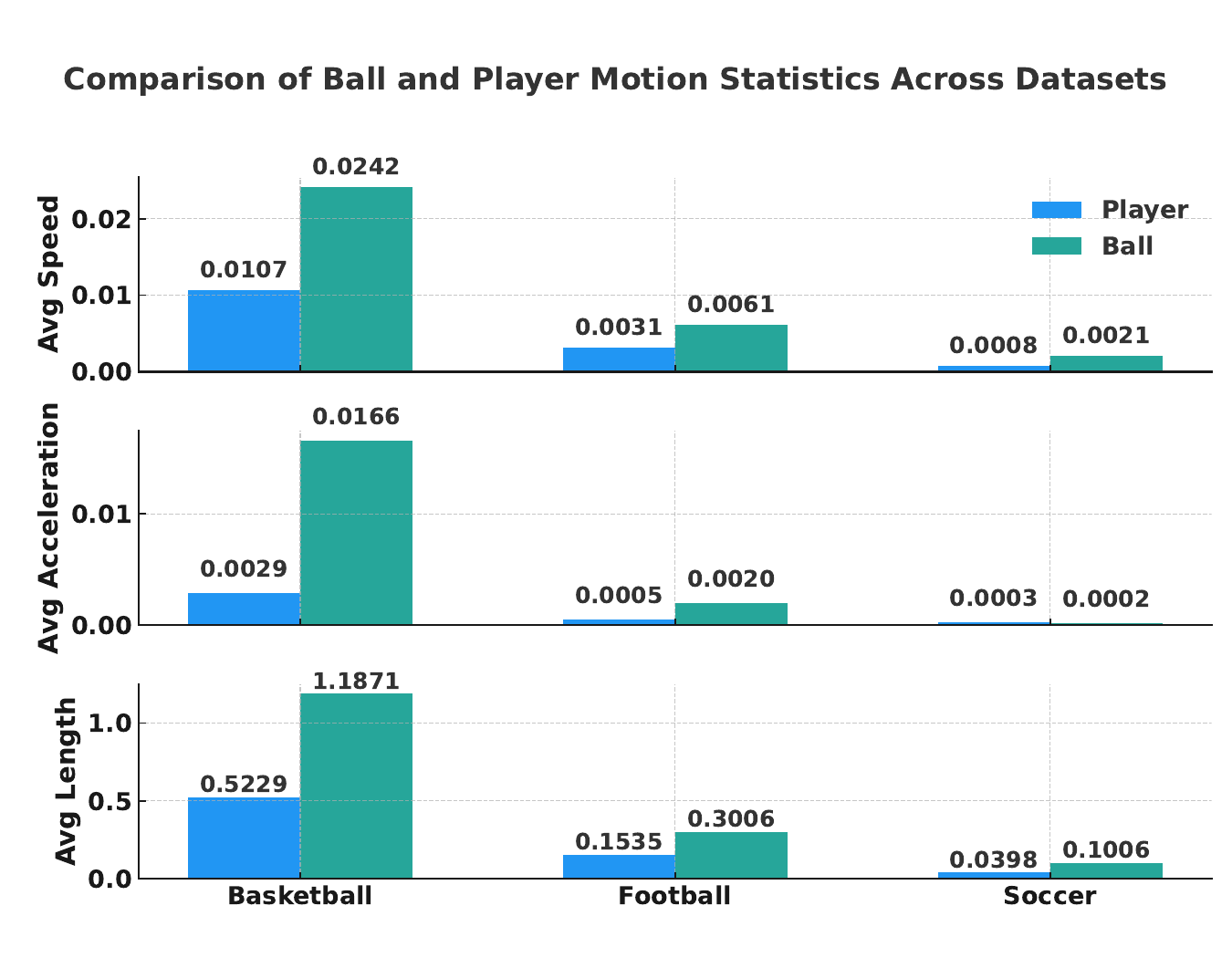}
\caption{Comparison of average speed, acceleration, and trajectory length for players and the ball in Basketball-U, Football-U, and Soccer-U datasets. The motion statistics reveal clear structural differences both between agent roles and across domains, motivating the need for adaptive and role-aware modeling in multi-agent trajectory generation.}
\label{fig:teaser}
\end{figure}

More recently, several works~\cite{xu2025sportstraj, capellera2025unified} have proposed unified frameworks capable of handling multiple trajectory modeling tasks, including prediction, imputation, and spatio-temporal recovery, within a general setting. New evaluation protocols and curated sports datasets have been introduced to support this unified formulation, often with strong baselines based on Conditional Variational Autoencoders (CVAEs). However, despite these advances, most methods continue to make domain-specific assumptions or treat all agents uniformly. They are typically trained and tested on the same dataset, overlooking two crucial challenges: (1) the heterogeneity of agent roles, where the ball exhibits drastically different motion patterns from human players; and (2) the distributional gap across sports domains, where different games have distinct spatial constraints, motion scales, and tactical styles. As a result, these models often fail to generalize across datasets, and naively combining multiple sports domains during training can lead to degraded performance due to latent conflicts in dynamics.

To quantitatively assess these challenges, we normalize trajectories from different sports datasets~\cite{xu2025sportstraj} into a common range of $[0, 1]$ and compute average speed, acceleration, and trajectory length per agent type. As illustrated in Fig.~\ref{fig:teaser}, there exists a clear and structured distributional gap across both roles and sports domains. The ball consistently exhibits higher velocity and acceleration than players, particularly in basketball, while movements in football and soccer are more constrained. Furthermore, the ball's behavior varies significantly across sports, indicating that both role-specific and domain-specific adaptation are essential for effective modeling. 

In addition to normalized comparisons, we also examine absolute motion statistics under original coordinate units to better understand the scale and semantics of each domain. We follow the metric definitions to compute the ground-truth statistics of player and ball motion in each dataset. As shown in Tab.~\ref{tab:gt_metrics}, we report the average values of step-wise displacement (Step), total path length (Path-L), and displacement from start to end (Path-D) across trajectories. The ball consistently exhibits higher Path-L and Path-D than players, indicating more jumpy and long-range movement compared to the smoother trajectories of players. In the soccer dataset, absolute values appear larger overall, however, this is due to the coordinate system being in pixels, which does not imply higher speed or motion magnitude compared to football (yards) or basketball (feet). The results reveal significant differences in motion magnitude between players and balls, as well as across datasets. These discrepancies further highlight the inherent distributional gaps between agent roles and sports domains, supporting our motivation for adaptive modeling.

\begin{table}[!t]
    \centering
    \caption{Ground-truth motion statistics for different agent roles (Player and Ball)  across the three sports datasets. Note that units differ across datasets: basketball is measured in feet, football in yards, and soccer in pixels.}
    \begin{tabular}{l|c|ccc}
    \toprule
     \multirow{2}{*}{\textbf{Dataset}}
     & \multirow{2}{*}{\textbf{Role}}
     & \multicolumn{3}{c}{\textbf{Ground-truth Metrics}} \\
     \cmidrule(lr){3-5}
     &  & \textbf{Step} & \textbf{Path-L} & \textbf{Path-D} \\
    \cmidrule(lr){1-5}
     \multirow{2}{*}{Basketball-U} & Player & 0.12 & 34.13 & 139.71 \\
     & Ball & 0.68 & 72.36 & 395.79 \\
     \cmidrule(lr){1-5}
     \multirow{2}{*}{Football-U} & Player & 0.02 & 11.95 & 51.43 \\
     & Ball & 0.14 & 26.03 & 146.39 \\
     \cmidrule(lr){1-5}
     \multirow{2}{*}{Soccer-U} & Player & 0.52 & 105.82 & 741.39 \\
     & Ball & 0.40 & 269.00 & 1015.34	\\
      \bottomrule
    \end{tabular}
    \label{tab:gt_metrics}
\end{table}

To overcome these challenges, we introduce \textbf{AdaSports-Traj}, an adaptive framework designed for multi-agent trajectory modeling in sports. The core of our approach is a Role- and Domain-Aware Adapter, which dynamically modulates latent representations based on both agent identity and domain context. This adapter is lightweight, plug-and-play, and compatible with a wide range of generative modeling backbones. 
Additionally, we design a Hierarchical Contrastive Learning objective, which separates supervision for role-sensitive and domain-aware alignment into two distinct embedding spaces. This strategy enables structural disentanglement while avoiding optimization conflict between contrastive signals.

We evaluate AdaSports-Traj on three sports datasets, Basketball-U, Football-U, and Soccer-U, showing that it consistently outperforms existing baselines in both within-domain and cross-domain settings. Ablation studies verify the contribution of the proposed adapter and contrastive loss in enabling role-aware and domain-adaptive modeling. The key contributions of our work can be summarized as follows:
\begin{itemize}
    \item We identify and quantify structured distributional gaps across agent roles and sports domains, and establish a strong unified trajectory modeling baseline under realistic multi-sport conditions.
    \item We propose AdaSports-Traj, an adaptive framework equipped with a Role- and Domain-Aware Adapter to dynamically modulate latent representations based on role and domain. 
    \item We introduce a Hierarchical Contrastive Learning strategy that supervises role-sensitive and domain-aware representations in separate subspaces, enabling structural disentanglement while mitigating optimization conflict.
    \item Comprehensive experiments confirm that the proposed framework not only achieves superior performance but also generalizes well across sports datasets and agent types, demonstrating strong robustness and adaptability.
\end{itemize}

%% file: section/3_relatedwork.tex
\section{Related Work}
\subsection{Trajectory Modeling and Completion}
Trajectory modeling is a fundamental approach for understanding multi-agent movement patterns. Despite challenges such as dynamic inter-agent interactions and incomplete observations, extensive efforts have been devoted to tasks like trajectory prediction and trajectory imputation.
In the domain of trajectory prediction, the goal is to forecast future agent movements conditioned on historical observations. 
A seminal method, Social-LSTM~\cite{alahi2016social}, introduced a pooling mechanism for agent interaction modeling. 
Follow-up works~\cite{hu2020collaborative, xu2021tra2tra} extended this idea to extract richer social features.
More recent studies~\cite{xu2022groupnet, li2022graph} leverage Graph Neural Networks (GNNs) to represent agents as nodes in an interaction graph. 
To capture the uncertainty and multimodality of future movements, generative models have gained popularity, including GANs~\cite{li2019conditional, amirian2019social}, CVAEs~\cite{salzmann2020trajectronplusplus, xu2022socialvae}, and diffusion-based models~\cite{gu2022stochastic, jiang2023motiondiffuser}.

In the domain of trajectory imputation, early approaches relied on statistical methods such as mean or median substitution~\cite{acuna2004treatment}, linear regression~\cite{ansleyestimation}, k-nearest neighbors~\cite{troyanskaya2001missing, beretta2016nearest}, and the EM algorithm~\cite{ghahramani1993supervised, nelwamondo2007missing}. 
These techniques often lack flexibility and struggle with complex dynamics. 
More recent works introduce deep learning-based imputation for sequential data, including autoregressive RNNs~\cite{cao2018brits} and generative models based on GANs, VAEs, or diffusion~\cite{qi2020imitative, miao2021generative, wen2024diffimpute, chen2024temporal, yuan2024diffusion}.
Yet, only a few works explicitly focus on multi-agent trajectory imputation. For instance, NAOMI~\cite{liu2019naomi} adopts a non-autoregressive approach, while GMAT~\cite{zhan2018generating} introduces hierarchical macro-intent modeling. Graph Imputer~\cite{omidshafiei2021time} combines forward-backward reasoning in soccer scenarios. More recent works bridge imputation and prediction: INAM~\cite{qi2020imitative} employs imitation learning to unify both tasks asynchronously, and GC-VRNN~\cite{xu2023uncovering} proposes a multi-task GCN-RNN framework to jointly perform prediction and imputation.

While these methods focus on reconstructing trajectories from partial observations, Traj-MAE~\cite{chen2023traj} introduces a continual pretraining framework for vehicle trajectory modeling by leveraging spatial maps. However, it is primarily limited to map-centric driving environments. In contrast, UniTraj~\cite{xu2025sportstraj} pioneers a general and unified trajectory modeling framework that supports diverse input conditions and tasks, with a focus on the sports domain, where social interactions are dense and structured. UniTraj establishes strong baselines across three sports datasets and handles various trajectory completion scenarios in a single model.
Nonetheless, UniTraj does not account for the substantial distributional gaps between different agent roles (e.g., ball vs. player) and across sports domains (e.g., basketball vs. soccer). This lack of structural adaptation limits its generalizability under domain shifts. In our work, we build upon this unified foundation and introduce a new perspective: explicitly modeling both intra-domain (role-level) and inter-domain (sports-level) distribution shifts. We propose a novel adaptive framework that incorporates structural awareness and improves generalization across agents and domains.

\subsection{Cross-Domain and Adaptive Trajectory Modeling}
Addressing domain shift in trajectory modeling is not new. Early works such as T-GNN~\cite{xu2022adaptive} and HATN~\cite{wang2022transferable} pioneered hierarchical frameworks to tackle this challenge. At a high level, these methods adopt a two-branch architecture, where one branch learns domain-invariant trajectory representations while the other captures domain-specific, transferable features. While effective, such methods typically rely on access to target-domain labels in the training phase, which limits their scalability. Additionally, they are not inherently online-adaptive, making them less suitable for real-time or open-world deployment.
Building upon this paradigm, follow-up studies~\cite{liu2024gate, li2024dual, loh2024cross, qian2025tracer, messaoud2025towards} further focus on aligning source- and target-domain representations at the latent level.
Popular approaches include domain adversarial training, maximum mean discrepancy (MMD), or contrastive learning to mitigate the distribution gap between domains. However, these techniques often assume access to both domains during training and can suffer when target domain distributions shift drastically.

Another line of research explores meta-learning~\cite{finn2017model}, which focuses on improving adaptation under limited data. In particular, few-shot and personalized trajectory prediction have been enhanced through meta-gradient updates~\cite{xu2022metaptp, shi2023metatraj, ivanovic2023expanding}. While meta-learning provides promising results, it usually requires prior exposure to the target domain during meta-training, restricting its general applicability in unseen environments. 

In contrast, we focus on a fully unified modeling scenario across diverse sports domains, where large distributional gaps exist both within each domain (e.g., ball vs. player) and across domains (e.g., basketball vs. soccer). Our method adapts to both intra-domain (roles) and inter-domain (sports) shifts without requiring target-domain labels, and achieves robust generalization through lightweight latent adaptation and hierarchical contrastive learning.

\subsection{Contrastive Learning for Trajectory Modeling}
Contrastive learning has proven to be a powerful method for learning discriminative and generalizable representations in self-supervised or weakly supervised settings. Initially popularized in visual representation learning~\cite{chen2020simple, he2020momentum}, it has since been extended to structured domains such as natural language~\cite{gao2021simcse} and time-series data~\cite{eldele2021time}.

In the context of multi-agent trajectory modeling, several recent works~\cite{kozerawski2022taming, zhou2022long, zhang2024tract} have leveraged contrastive objectives to address the long-tail distribution problem commonly observed in trajectory prediction. Specifically, methods such as~\cite{makansi2021exposing, wang2023fend} apply contrastive loss over implicitly clustered trajectory classes to separate rare trajectories from common ones in the embedding space.
While prior works primarily focus on semantic diversity across trajectory types, our method introduces a hierarchical contrastive strategy that explicitly disentangles role-specific and domain-specific information, improving generalization across agents and sports.

%% file: section/4_method.tex
\section{Method}
\subsection{Problem Formulation}
We address the general problem of trajectory modeling under arbitrary observation conditions, following the definition first introduced in prior work~\cite{xu2025sportstraj}.
In this formulation, any input trajectory sequence is treated as a partially observed sequence, where the known portions serve as constraints (i.e., inputs), and the unknown portions constitute the prediction targets.
Formally, let the complete trajectory be denoted as $X \in \mathbb{R}^{N\times T \times D}$, where $N$ is the number of agents, $T$ is the trajectory length, and $D$ is the spatial dimension of each agent's state (typically $D=2$ for 2D positions).
The state of agent $i$ at time step $t$ is denoted as $\boldsymbol{x}^{t}_{i} \in \mathbb{R}^{D}$.
To model missing observations, we define a binary mask matrix $M \in \mathbb{R}^{N\times T}$ valued in $\{0, 1\}$, where $m^{t}_{i}=0$ indicates it is missing. We assume that when an agent is unobserved, all spatial dimensions are missing simultaneously.
The trajectory can thus be partitioned into two parts: the visible (observed) region $X_{v} = X \odot M$ and the missing region $X_{m} = X \odot (\mathbf{1} - M)$. 
The objective is to recover the full trajectory by generating both reconstructed and missing portions, denoted by $\hat{Y} = \{\hat{X}_{v}, \hat{X}_{m}\}$.
The original ground-truth trajectory is denoted as $Y = X = \{X_{v}, X_{m}\}$.
The goal is to learn a generative model $f(\cdot)$ parameterized by $\theta$, which generates a complete trajectory $\hat{Y}$.
A widely adopted method for parameter estimation is to factorize the joint trajectory distribution and maximize the log-likelihood:
\begin{equation}
\begin{aligned}
    \theta^{*} &= \arg \max_{\theta} \sum_{\boldsymbol{x}^{\leq T} \in \Omega} \log p_{\theta}(\boldsymbol{x}^{\leq T}) \\
&= \arg \max_{\theta} \sum_{\boldsymbol{x}^{\leq T} \in \Omega} \sum_{t=1}^T \log p_{\theta}(\boldsymbol{x}^{t} | \boldsymbol{x}^{<t}),
\end{aligned}
\label{eq:loglike}
\end{equation}
where $\Omega = \{1, 2, ..., N\}$ denotes the agent set, and $\boldsymbol{x}^{\leq T}$ denotes the sequential trajectory of the agent.

\subsection{Conditional Variational Trajectory Modeling}
To address the general trajectory modeling problem, many prior works adopt a generative framework, which enables flexible handling of uncertainty and partial observability. In our work, we follow this generative formulation while also incorporating evaluation protocols that are specifically designed to reveal distributional gaps both within and across different sports domains.
Formally, the generative model parameters $\theta$ are optimized by maximizing the sequential evidence lower bound (ELBO) as follows:
\begin{equation}
\begin{aligned}
\mathbb{E}_{q_{\phi}(\boldsymbol{z}^{\leq T} | \boldsymbol{x}^{\leq T})}
& \left[\sum^{T}_{t=1}\log p_{\theta}(\boldsymbol{x}^{t} | \boldsymbol{z}^{\leq t}, \boldsymbol{x}^{<t}) \right. \\
& \left.-\text{KL}\left(q_{\phi}(\boldsymbol{z}^{t} | \boldsymbol{x}^{\leq t}, \boldsymbol{z}^{<t}) \|
p_{\theta}(\boldsymbol{z}^{t} | \boldsymbol{x}^{<t}, \boldsymbol{z}^{<t})\right)\right]
\label{eq:elbo}
\end{aligned}
,
\end{equation}
where $\boldsymbol{z}$ denotes the latent variables across all agents. The prior $p_{\theta}(\boldsymbol{z}^{t} | \boldsymbol{x}^{<t}, \boldsymbol{z}^{<t})$ is typically modeled as a Gaussian distribution. The encoder $q_{\phi}(\boldsymbol{z}^{t} | \boldsymbol{x}^{\leq t}, \boldsymbol{z}^{<t})$ approximates the posterior, while the decoder $ p_{\theta}(\boldsymbol{x}^{t} | \boldsymbol{z}^{\leq t}, \boldsymbol{x}^{<t})$ reconstructs the trajectory given latent and observed variables. This formulation serves as a variational lower bound of the log-likelihood defined in~\eqref{eq:loglike}, computed across each timestep.

\begin{figure}[!t]
  \centering
   \includegraphics[width=0.9\linewidth]{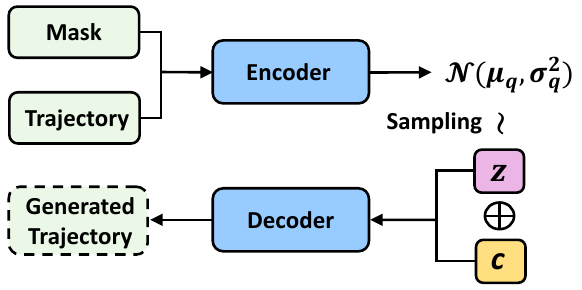}
   \caption{Overview of the UniTraj~\cite{xu2025sportstraj} base framework, which adopts a CVAE structure. The encoder maps the masked trajectory into latent distribution parameters, and the decoder produces the complete trajectory conditioned on both the sampled latent variable and the conditional features.}
   \label{fig:unitraj}
\end{figure}

In our work, we adopt the pioneering method UniTraj~\cite{xu2025sportstraj} as our base framework. 
As presented in Fig.~\ref{fig:unitraj}, UniTraj comprises a spatio-temporal encoder and a decoder.
The encoder transforms the observed trajectories into posterior parameters of a Gaussian latent distribution. Samples from this distribution are then combined with condition features and fed into the decoder to reconstruct or generate the complete trajectory. The latent variables thus act as compact feature summaries that encode the underlying dynamics and uncertainties.

However, existing methods typically train and evaluate models on a single dataset, without accounting for potential distributional gaps across roles or domains. As a result, the latent space fails to reflect these structured discrepancies. In contrast, our insight is to enhance the latent representation by explicitly disentangling variations across roles and domains. This disentanglement encourages the latent space to encode meaningful structural differences and better capture the distributional shifts present in diverse sports.

\begin{figure*}[!t]
  \centering
   \includegraphics[width=0.95\linewidth]{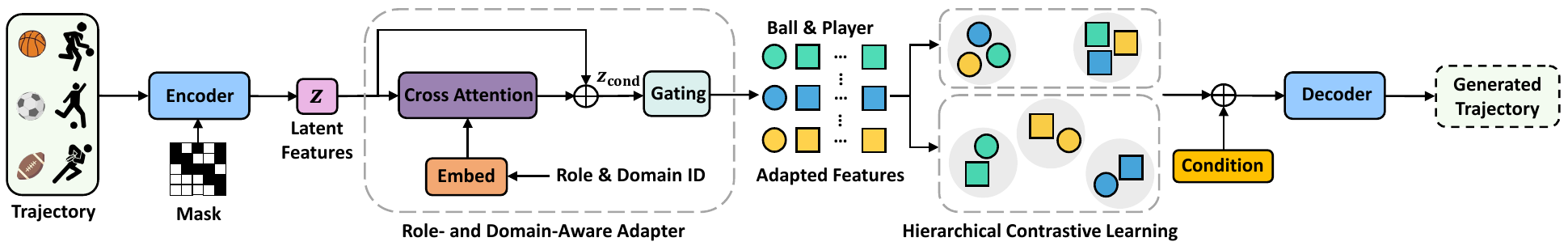}
   \caption{Overall architecture of our AdaSports-Traj. Masked trajectories are encoded and adaptively modulated by a role- and domain-aware adapter. Hierarchical contrastive learning further disentangles structure in the latent space before decoding the full trajectory.}
   \label{fig:framework}
\end{figure*}

\subsection{Role- and Domain-Aware Adapter}
In multi-agent sports scenarios, agents exhibit distinct motion patterns not only due to their semantic roles (e.g., ball vs. player) but also due to the specific domain context (e.g., basketball vs. soccer). To enable adaptive modulation of latent features based on both factors, we propose a modular Role- and Domain-Aware Adapter.

Each agent is associated with a discrete role label $r$ and a shared domain label $d$. These labels are first embedded into fixed-dimensional vectors as:
\begin{equation}
    \begin{aligned}
        \boldsymbol{e}_{\text{role}} = \text{Embedding}(r) \\
        \boldsymbol{e}_{\text{domain}} = \text{Embedding}(d) \\
    \end{aligned}
    ,
\end{equation}
where $\text{Embedding}(\cdot)$ denotes a learnable embedding layer. The embeddings are summed to form a conditioning token that encodes both semantic and contextual priors for each agent.

We then perform multi-head cross-attention, using this conditioning token as the query, and the original latent features $\boldsymbol{z}$ as both key and value, defined as:
\begin{equation}
\begin{aligned}
        query &= \boldsymbol{e}_{\text{role}}+\boldsymbol{e}_{\text{domain}} \\
        key  &= value = \boldsymbol{z} \\
        \boldsymbol{z}_{\text{cond}} &= \text{CrossAttention}(query, key, value)
\end{aligned}
.
\end{equation}

To preserve the original latent representation while introducing adaptive bias, we employ a token-wise gating mechanism. Specifically, we compute a gating weight $\alpha_{\text{gate}}$ by applying a sigmoid-activated MLP to the concatenation of the original feature $\boldsymbol{z}$ and the conditioned feature $\boldsymbol{z}_{\text{cond}}$. The adapted representation is then obtained by interpolating between the two inputs $\boldsymbol{z}$ and $\boldsymbol{z}_{\text{cond}}$ in an element-wise manner:
\begin{equation}
    \begin{aligned}
        \alpha_{\text{gate}} &= \text{Sigmoid}(\text{MLP}([\boldsymbol{z}, \boldsymbol{z}_{\text{cond}}])) \\
        \boldsymbol{z}_{\text{adapted}} &=\alpha_{\text{gate}} \odot \boldsymbol{z}_{\text{cond}} + (1 - \alpha_{\text{gate}})\odot \boldsymbol{z}
    \end{aligned}
    ,
\end{equation}
where $[\boldsymbol{z}, \boldsymbol{z}_{\text{cond}}]$ denotes feature concatenation and $\odot$ represents element-wise multiplication.
This strategy allows the adapter to softly balance between domain and role-informed updates and the original representation, enabling fine-grained control over information integration at the agent level.

Overall, the proposed adapter supports dynamic, structure-aware feature adaptation, enabling latent representations to specialize according to both semantic role and domain context. 
It can be flexibly inserted at various stages of the generative process, such as after the encoder, prior, or posterior, without introducing significant computational overhead. Its plug-and-play design allows seamless integration into CVAE-based or diffusion-based trajectory models, enhancing their ability to specialize latent representations for diverse structural contexts and improving generalization across heterogeneous agent configurations and sports domains.

\subsection{Hierarchical Contrastive Learning}
To further enhance structure-aware representation learning, we propose a Hierarchical Contrastive Learning objective that explicitly separates supervision along two orthogonal axes: agent roles and sports domains. Instead of applying a unified contrastive loss on the entire latent space, which can lead to conflicting optimization signals, we employ two dedicated projection heads that map the adapted latent features into role-sensitive and domain-sensitive subspaces, respectively. This design allows the model to achieve fine-grained structural alignment while preserving representational disentanglement.

Let $\boldsymbol{z}_{\text{adapted}}$ denote the adapted latent representation obtained from the adapter. We project this representation into two distinct spaces via two parallel MLPs:
\begin{equation}
    \begin{aligned}
        \boldsymbol{z}_\text{role} &= \text{MLP}_{\text{adapted}}(\boldsymbol{z}_{\text{adapted}}) \\
        \boldsymbol{z}_\text{domain} & = \text{MLP}_{\text{adapted}}(\boldsymbol{z}_{\text{adapted}})
    \end{aligned}
    .
    \label{eq:proj}
\end{equation}

Each projection is followed by $L_{2}$ normalization, and a standard InfoNCE~\cite{oord2018representation} loss is applied within each subspace.
The role contrastive loss encourages agents with the same role (e.g., ball or player), regardless of domain, to be close in the role-sensitive space. In contrast, the domain contrastive loss aligns agents within the same sports domain (e.g., basketball), regardless of their semantic roles.
Formally, for a positive pair $(i, j)$, the losses are computed as:
\begin{equation}
    \begin{aligned}
        \mathcal{L}_{\text{role}} &= -\log\frac{\exp(\text{sim}(\boldsymbol{z}_{i}^{{\text{role}}}, \boldsymbol{z}_{j}^{{\text{role}}})/\tau_{c})}{\sum_{k\neq i}\exp(\text{sim}(\boldsymbol{z}_{i}^{{\text{role}}}, \boldsymbol{z}_{j}^{{\text{role}}})/\tau_{c})} \\
        \mathcal{L}_{\text{domain}} &= -\log\frac{\exp(\text{sim}(\boldsymbol{z}_{i}^{\text{domain}}, \boldsymbol{z}_{j}^{\text{domain}})/\tau_{c})}{\sum_{k\neq i}\exp(\text{sim}(\boldsymbol{z}_{i}^{\text{domain}}, \boldsymbol{z}_{j}^{\text{domain}})/\tau_{c})}
    \end{aligned}
,
\label{eq:contrastive}
\end{equation}
where $\text{sim}(\cdot,\cdot)$ denotes cosine similarity and 
$\tau_{c}$ is a temperature hyperparameter. The positive pair
$(i,j)$ is selected either by shared role or shared domain, depending on the objective.

The overall hierarchical contrastive loss combines both as:
\begin{equation}
    \mathcal{L}_{\text{hier}} = \mathcal{L}_{\text{role}} + \lambda_{c}\mathcal{L}_{\text{domain}},
\end{equation}
where $\lambda_{c}$ balances the contribution of each contrastive term.

This hierarchical design promotes the model’s ability to learn disentangled and generalizable latent representations while avoiding optimization conflicts that may arise when jointly optimizing for multiple alignment criteria in a single space.
    
\subsection{Overall Loss Function}
Following the baseline framework, the model is trained using a combination of three core loss terms: (1) ELBO loss for latent modeling, (2) reconstruction loss on the visible regions, and (3) Winner-Take-All (WTA) loss over the $K$ generated trajectory samples:
\begin{equation}
    \begin{aligned}
        \mathcal{L}_{\text{elbo}} &= \|\hat{X}_{m} - {X}_{m}\|^2_2 +   \lambda_{1}\text{KL}\left(\mathcal{N}(\boldsymbol{\mu}_{q}, \text{Diag}(\boldsymbol{\sigma}_{q}^2)) \| \mathcal{N} (0, \mathbf{I})
        \right) \\
         \mathcal{L}_{\text{rec}} & = \|\hat{X}_{v} - {X}_{v}\|^2_2 \\
        \mathcal{L}_{\text{wta}} & =  \min_{K}\|\hat{Y}^{(k)} - Y\|^2_2
    \end{aligned}
    ,
    \label{eq:loss}
\end{equation}
where $\hat{X}_{m}$ and ${X}_{m}$ denote the generated and ground-truth missing regions, respectively, while $\hat{X}_{v}$ and ${X}_{v}$ denote the visible regions. The WTA loss encourages sample diversity by selecting the best-matching trajectory among $K$ generated trajectories $\hat{Y}^{(k)}$.

In addition to these baseline objectives, we incorporate our proposed hierarchical contrastive loss $\mathcal{L}_{\text{hier}}$, which enhances latent structure alignment across roles and domains. The final training objective combines all terms as follows:
\begin{equation}
    \mathcal{L} = \mathcal{L}_{\text{elbo}} + \lambda_{2}\mathcal{L}_{\text{rec}} + \lambda_{3}\mathcal{L}_{\text{wta}} + \lambda_{4}\mathcal{L}_{\text{hier}},
\end{equation}
where $\lambda_{1}$, $\lambda_{2}$, $\lambda_{3}$, and $\lambda_{4}$ are weighting hyperparameters that balance the contributions of each loss component.

%% file: section/5_experiments.tex
\section{Experiments}
\input{table/S2S_basketball_football}
\input{table/S2S_soccer}
\subsection{Benchmarks and Setup}
\subsubsection{Datasets}
We evaluate our method on three sports datasets introduced in~\cite{xu2025sportstraj}.
\textbf{(1) Basketball-U} consists of 93,490 training samples and 11,543 testing samples, where each sequence records trajectories of 1 ball, 5 offensive players, and 5 defensive players.
\textbf{(2) Football-U} provides 10,762 training samples and 2,624 testing samples, each covering trajectories for 1 ball and 22 players (11 per team).
\textbf{(3) Soccer-U} includes 9,882 training samples and 2,448 testing samples, with trajectories for 1 ball and 22 players (11 offensive and 11 defensive) per sequence.
All datasets follow the unified multi-agent trajectory format, where agents’ positions are recorded over time. These datasets span multiple sports with distinct dynamics, making them well-suited for evaluating role- and domain-adaptive trajectory modeling.

\subsubsection{Evaluation Protocol}
We consider two evaluation settings to assess both in-domain performance and cross-domain generalization capabilities:
\textbf{(1) Single-to-Single (S2S)}: In this standard setting, models are trained and evaluated independently on each dataset. That is, training and testing are performed within the same domain (e.g., training on Basketball-U and testing on Basketball-U). This serves as a strong per-domain baseline and reflects performance under fully domain-specific supervision.
\textbf{(2) Unified-to-Single (U2S)}: To evaluate cross-domain generalization, we propose a more challenging setting where models are trained on a unified dataset \textbf{Sports-U}, constructed by merging the training sets of Basketball-U, Football-U, and Soccer-U, resulting in 114,134 total training sequences. The model is then tested separately on the test sets of the three original datasets. This setting requires the model to generalize to multiple sports with diverse dynamics and spatial layouts, exposing potential distributional gaps across domains.

\subsubsection{Metrics}
We use five evaluation metrics to assess performance:
\textbf{(1) minADE$_{20}$}: Minimum Average Displacement Error across 20 sampled trajectories. It measures the closest match between predictions and the ground truth.
\textbf{(2) Out-of-Boundary (OOB)}: Percentage of predicted coordinates that fall outside the valid spatial boundaries of the playing field.
\textbf{(3) Step}: Average step-wise displacement, capturing local movement smoothness
\textbf{(4) Path-L}: Average total trajectory length per agent, reflecting movement scale.
\textbf{(5) Path-D}: Maximum discrepancy in trajectory length among agents within the same sequence.
For minADE$_{20}$ and OOB, lower values indicate better performance. For Step, Path-L, and Path-D, performance is better when the values are closer to the ground truth.

\subsubsection{Baselines}
In the \textbf{S2S} setting, each baseline is trained and tested independently on each dataset. We evaluate our method against a diverse set of baselines across three categories:
\textbf{(1) Statistical Approaches:} These include simple but strong non-learning baselines such as Mean, Median, and Linear Fit, which extrapolate trajectories based on basic statistical properties of the observed data.
\textbf{(2) Vanilla Models:} We include standard deep sequence models such as LSTM~\cite{hochreiter1997long} and Transformer~\cite{vaswani2017attention}, which serve as foundational architectures for temporal modeling.
\textbf{(3) Advanced Learning-Based Baselines:} We compare our approach against state-of-the-art trajectory modeling methods, including: MAT~\cite{zhan2018generating}, Naomi~\cite{liu2019naomi}, INAM~\cite{qi2020imitative}, SSSD~\cite{alcaraz2022diffusion}, GC-VRNN~\cite{xu2023uncovering}, and UniTraj~\cite{xu2025sportstraj}, which represent a range of autoregressive and diffusion-based approaches for trajectory generation.

In the \textbf{U2S} setting, we use UniTraj~\cite{xu2025sportstraj} as the primary comparison baseline, as it is the pioneering method for unified trajectory modeling under a task-general framework. It provides a strong and structurally consistent baseline for evaluating generalization across multiple sports domains.

\subsubsection{Implementation Details}
For the S2S setting, we integrate our proposed approach into the UniTraj baseline by assigning a fixed domain label $d$ and applying only the role contrastive loss as defined in \eqref{eq:contrastive}.
For the U2S setting, we adopt the same encoder and decoder architecture as in the original UniTraj to ensure a fair comparison.
In both settings, we follow the model hyperparameter configurations used in UniTraj. Specifically, we set $\lambda_{c} = 1$ and $\lambda_{4} = 0.1$ to balance the hierarchical contrastive loss terms.
All experiments are conducted using PyTorch~\cite{paszke2019pytorch} on a NVIDIA A6000 GPU.
All models are trained for 100 epochs with a batch size of 128 using the Adam optimizer~\cite{diederik2015adam}.
Training takes approximately 24 hours for the U2S setting and between 8 and 15 hours for the S2S setting.
The learning rate is initialized at 0.001 and decays by 0.9 every 20 epochs.

\subsection{Main Results}
\input{table/U2S_basketball_football_soccer}
\input{table/ablation_module}
\subsubsection{S2S Performance}
Tab.~\ref{tab:S2S_basketball_football} and~\ref{tab:S2S_soccer} show the results of AdaSports-Traj and various baselines under the \textbf{S2S} setting.
Across all three datasets, our method demonstrates consistent improvements compared to the baselines.
On Basketball-U, it achieves a lower minADE$_{20}$ of 4.21 compared to 4.77 from UniTraj, producing more realistic motion patterns in Step and Path-L.
On Football-U, our method reaches the best minADE$_{20}$ of 3.04 and records the most human-like motion dynamics with the lowest Path-D.
Even on the more challenging Soccer-U dataset, where trajectories are recorded in pixel space, AdaSports-Traj achieves the best overall performance, reaching a minADE$_{20}$ of 91.52 and demonstrating enhanced realism across spatial metrics.
These results highlight that our method not only enhances prediction accuracy but also better captures the distinctive motion characteristics of each sport.

\subsubsection{U2S Performance}
Tab.~\ref{tab:U2S_all} presents the evaluation results under the U2S setting.
Compared to the S2S setting, we observe a notable performance degradation for the baseline UniTraj, highlighting its difficulty in generalizing under cross-domain distribution shifts.
Notably, the decline is especially pronounced in the Path-L and Path-D metrics, which are sensitive to differences in motion range and scale across sports. In contrast, our method consistently outperforms UniTraj across all three sports domains, demonstrating its effectiveness in addressing both intra- and inter-domain distribution gaps.

\input{table/ablation_adapter}
\input{table/ablation_contrastive_learning}

\subsection{Ablation Study}
\subsubsection{Contribution of RDA and HC}
We conduct an ablation study to
evaluate the effectiveness of each component within the AdaSports-Traj framework. 
As shown in Tab.~\ref{tab:ablation_module}, removing either the Role- and Domain-Aware Adapter (RDA) or the Hierarchical Contrastive (HC) loss leads to performance degradation across all datasets and settings. Notably, combining both components yields the best performance, demonstrating their complementary benefits for learning robust and generalizable trajectory representations.

\begin{figure*}[t]
  \centering
  \includegraphics[width=0.98\linewidth]{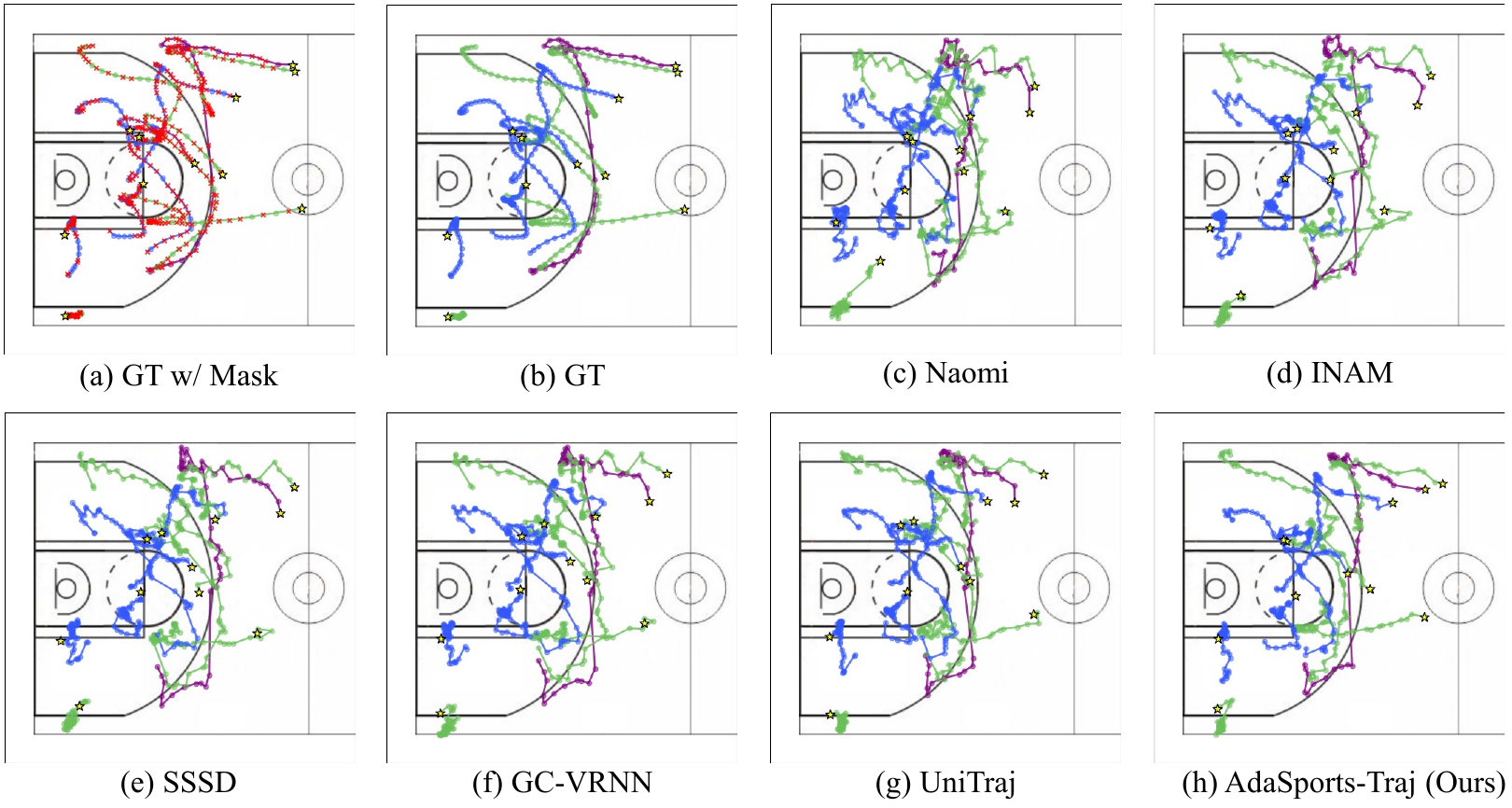}
   \caption{Qualitative comparison of baselines and our method. Ball trajectories are shown in purple, offensive players in green, and defensive players in blue. Red ``x'' markers denote masked (missing) locations, while yellow stars indicate starting positions.}
   \vspace{-3mm}
   \label{fig:vis_1}
\end{figure*}

\begin{figure}[t]
    \centering
    \includegraphics[width=0.48\linewidth]{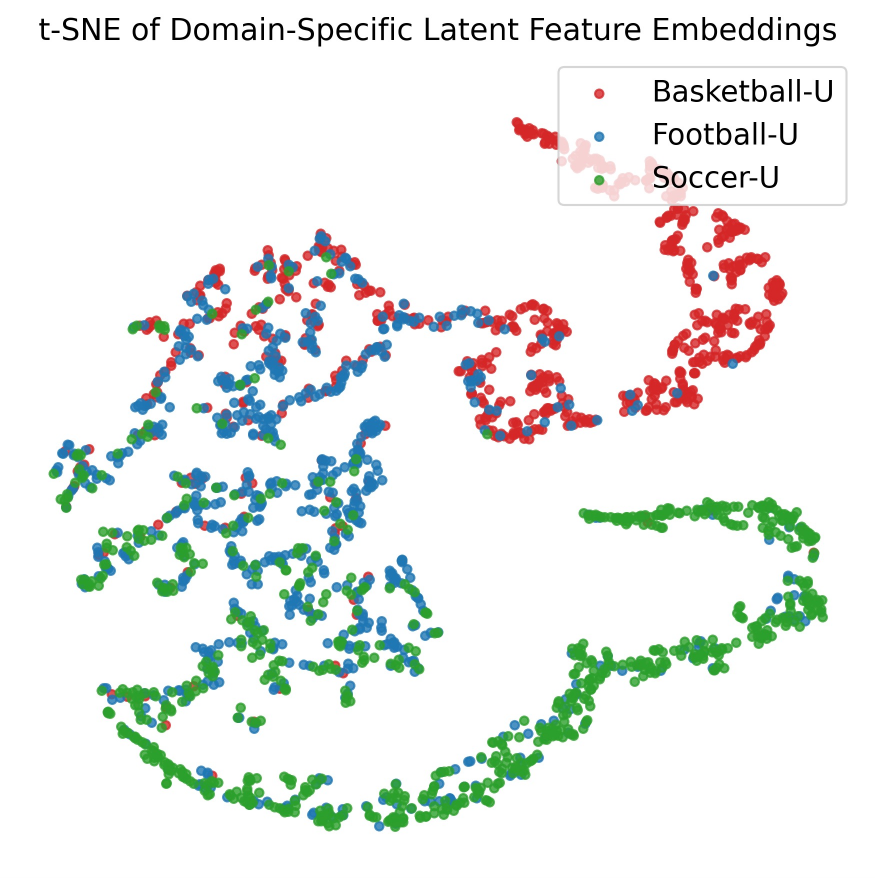}
    \includegraphics[width=0.48\linewidth]{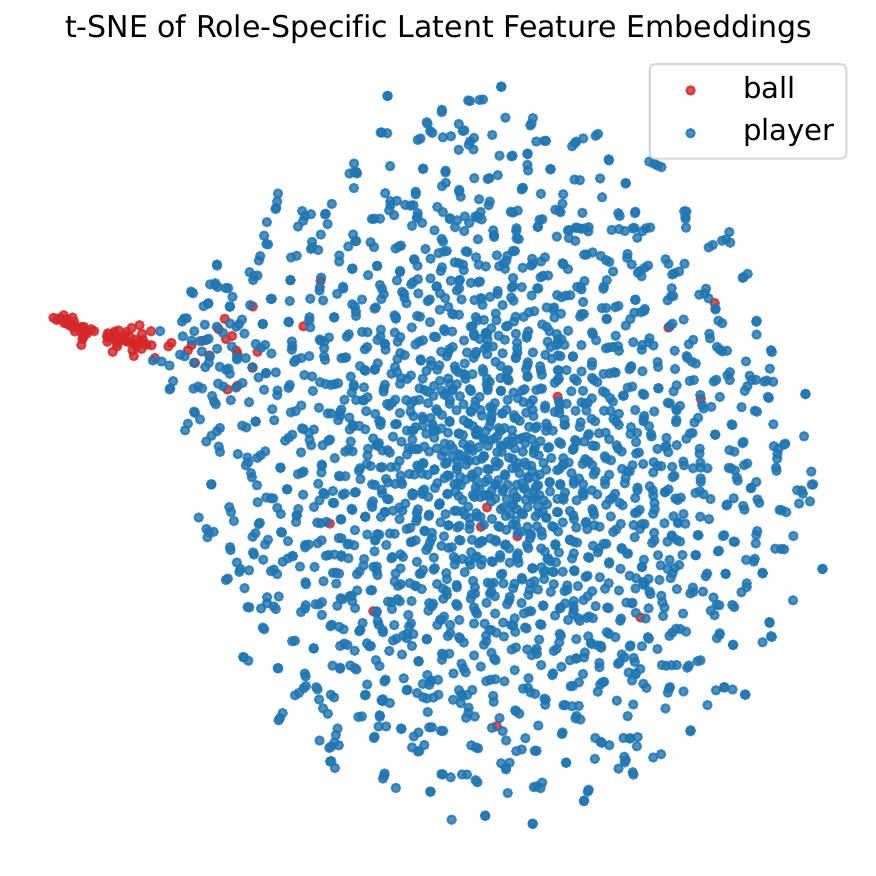}
    \caption{t-SNE visualizations of (left) domain-specific features and (right) role-specific features. Clear clustering patterns demonstrate our model's ability to learn semantically meaningful representations.}
    \vspace{-3mm}
    \label{fig:tsne_latent}
\end{figure}

\subsubsection{Adapter Designs}
We further investigate the impact of different adapter designs under both S2S and U2S settings. Specifically, we evaluate three variants: (1) ``w/o gating'': Remove the gating mechanism and directly use the attended latent $\boldsymbol{z}_{\text{cond}}$ as the adapted feature. (2) ``w/ feature-wise'': Apply a gating mechanism without the Sigmoid activation, where each feature dimension is modulated independently. (3) ``w/ token-wise'': Our proposed design that uses token-level gating weights learned via a Sigmoid-activated MLP to softly fuse $\boldsymbol{z}$ and $\boldsymbol{z}_{\text{cond}}$.
The results in Tab.~\ref{tab:ablation_adapter} show that our token-wise gating consistently achieves the best performance across all datasets and settings, emphasizing the importance of adaptive fusion at the token level.
The improvements are particularly substantial in the U2S setting, where domain-level shifts exist and the adapter must handle both role- and domain-specific variations. Interestingly, the feature-wise gating variant performs worse than simply removing the gating mechanism, possibly due to its limited capacity to preserve temporal and agent-level structure, which token-wise gating better retains.

\subsubsection{Contrastive Learning Design}
We introduce hierarchical contrastive learning that uses two independent projection heads to map the adapted latent features into separate role-sensitive and domain-sensitive subspaces. The key motivation is to prevent conflicting learning signals: role contrastive loss encourages separation between balls and players, while domain contrastive loss aligns agents from the same sports domain, regardless of role. 
To quantitatively assess this, we include a shared-features (direct add) variant, which removes the projection heads and applies both contrastive losses directly on the same latent features $\boldsymbol{z}_{\text{adapted}}$. As shown in Tab.~\ref{tab:ablation_contrastive}, this approach leads to significantly worse performance, validating the conflict between the two objectives when not disentangled in the latent space.
Additionally, both role-only and domain-only variants improve upon the no-contrastive baseline, indicating that each perspective contributes useful supervision. 
Notably, our full hierarchical design achieves the best results across all datasets, confirming the effectiveness of decoupled supervision for learning structurally aware representations.

\subsection{Qualitative Results}
\subsubsection{t-SNE of Learned Representations}
To qualitatively assess the learned representations, we visualize the projected domain- and role-specific latent features using t-SNE~\cite{van2008visualizing}, as defined by $\boldsymbol{z}_\text{domain}$ and $\boldsymbol{z}_\text{role}$ in \eqref{eq:proj}. Each point corresponds to the final-step representation of an agent, with colors indicating its domain or role (Fig.~\ref{fig:tsne_latent}).
The left subfigure shows clear clustering by domain: features from Basketball-U, Football-U, and Soccer-U form well-separated groups. This indicates that our model successfully captures domain-aware structure in the latent space, which is critical in the U2S setting where training and testing domains differ.
In the right subfigure, role-specific features exhibit a distinct separation between ball and player representations. The ball embeddings cluster tightly, reflecting their consistent functional role across scenarios, while player embeddings are more dispersed due to their greater role diversity.
These qualitative results demonstrate that our model captures semantically meaningful and structurally disentangled latent representations, enabling more robust generalization across both domain and role dimensions.

\subsubsection{Trajectory Visualization}
We provide qualitative comparisons under the S2S setting in Fig.~\ref{fig:vis_1}, where our generated trajectories are visibly closer to the ground truth, especially in regions with missing observations.
At masked positions denoted by red markers, our approach generates predictions with higher accuracy and improved contextual consistency. 
Additionally, our predicted trajectories are smoother than those of other baselines, highlighting the effectiveness of our adaptive design.

\begin{figure}[t]
    \centering
    \begin{subfigure}{0.32\linewidth}
        \includegraphics[width=\linewidth]{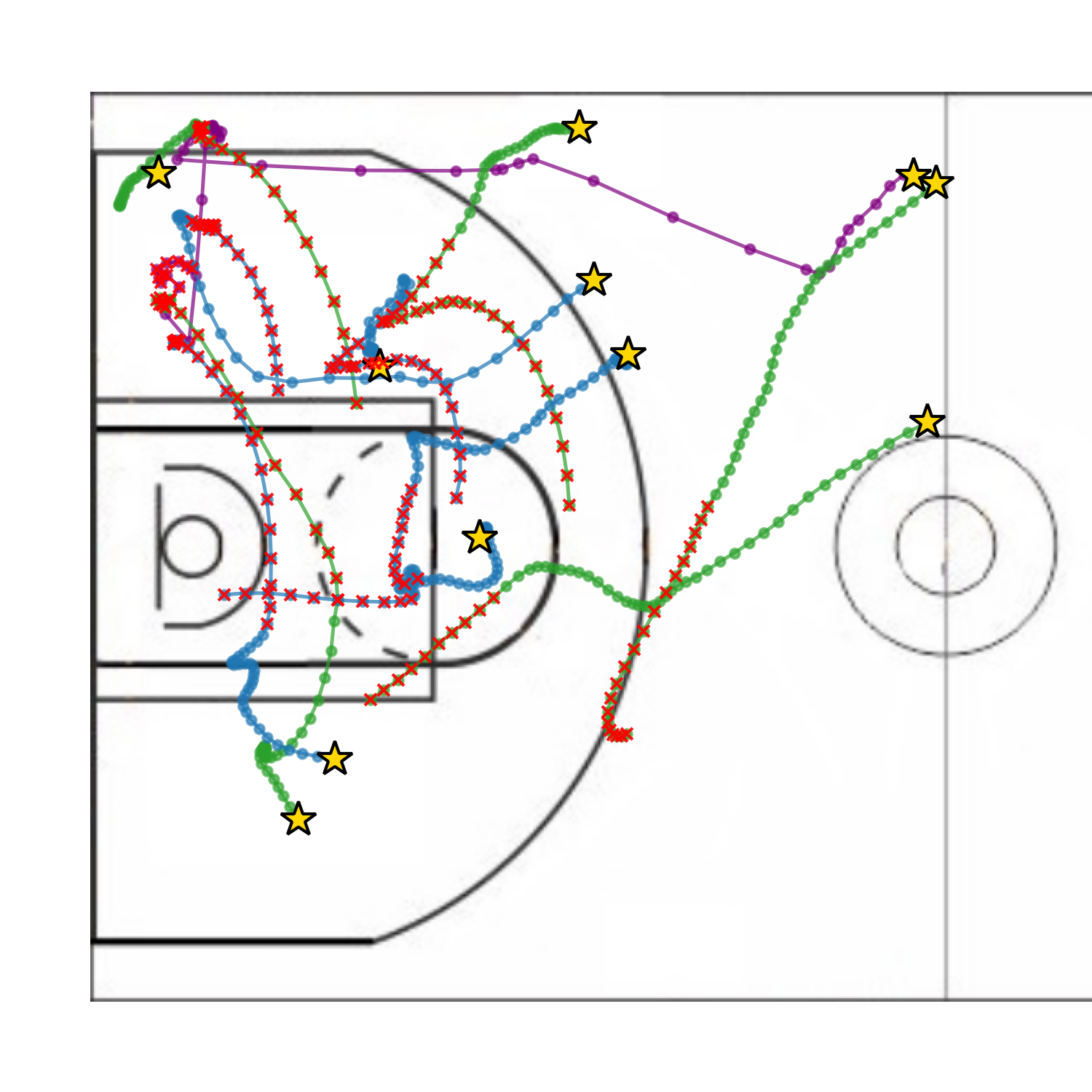}
        \caption{GT w/ Mask}
    \end{subfigure}
    \hfill
    \begin{subfigure}{0.32\linewidth}
        \includegraphics[width=\linewidth]{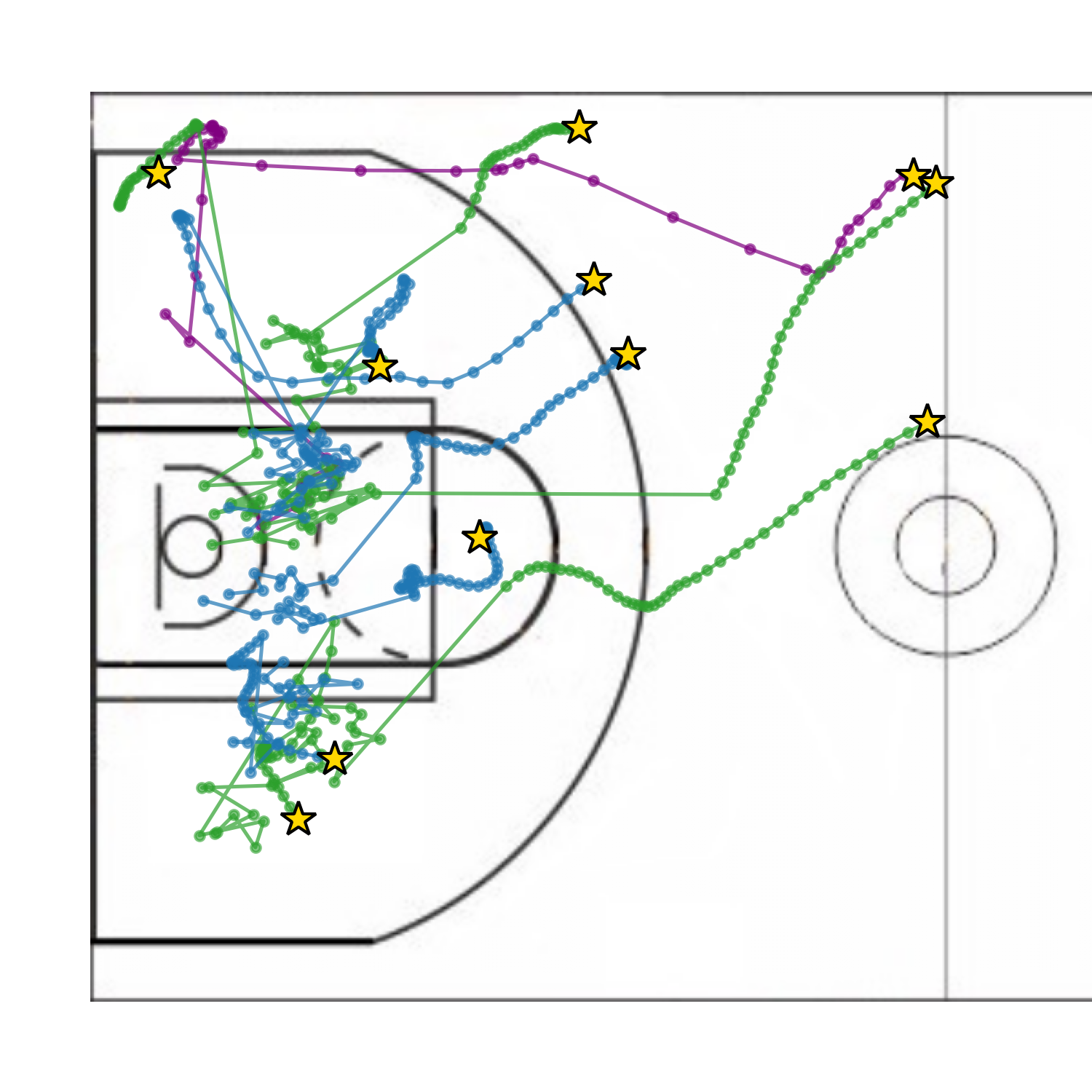}
        \caption{UniTraj}
    \end{subfigure}
    \hfill
    \begin{subfigure}{0.32\linewidth}
        \includegraphics[width=\linewidth]{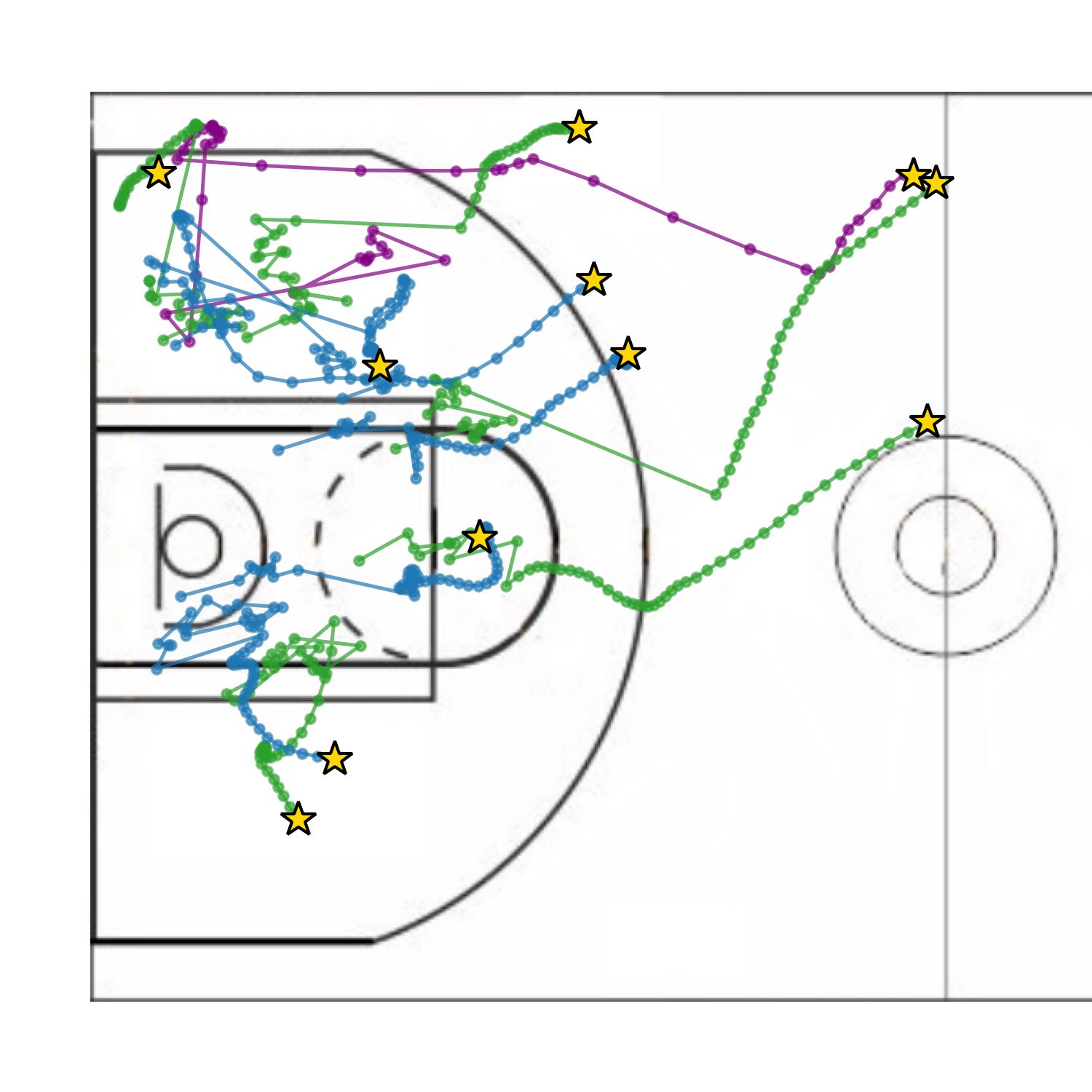}
        \caption{Ours}
    \end{subfigure}

    \vspace{-0.5mm} 

    \begin{subfigure}{0.32\linewidth}
        \includegraphics[width=\linewidth]{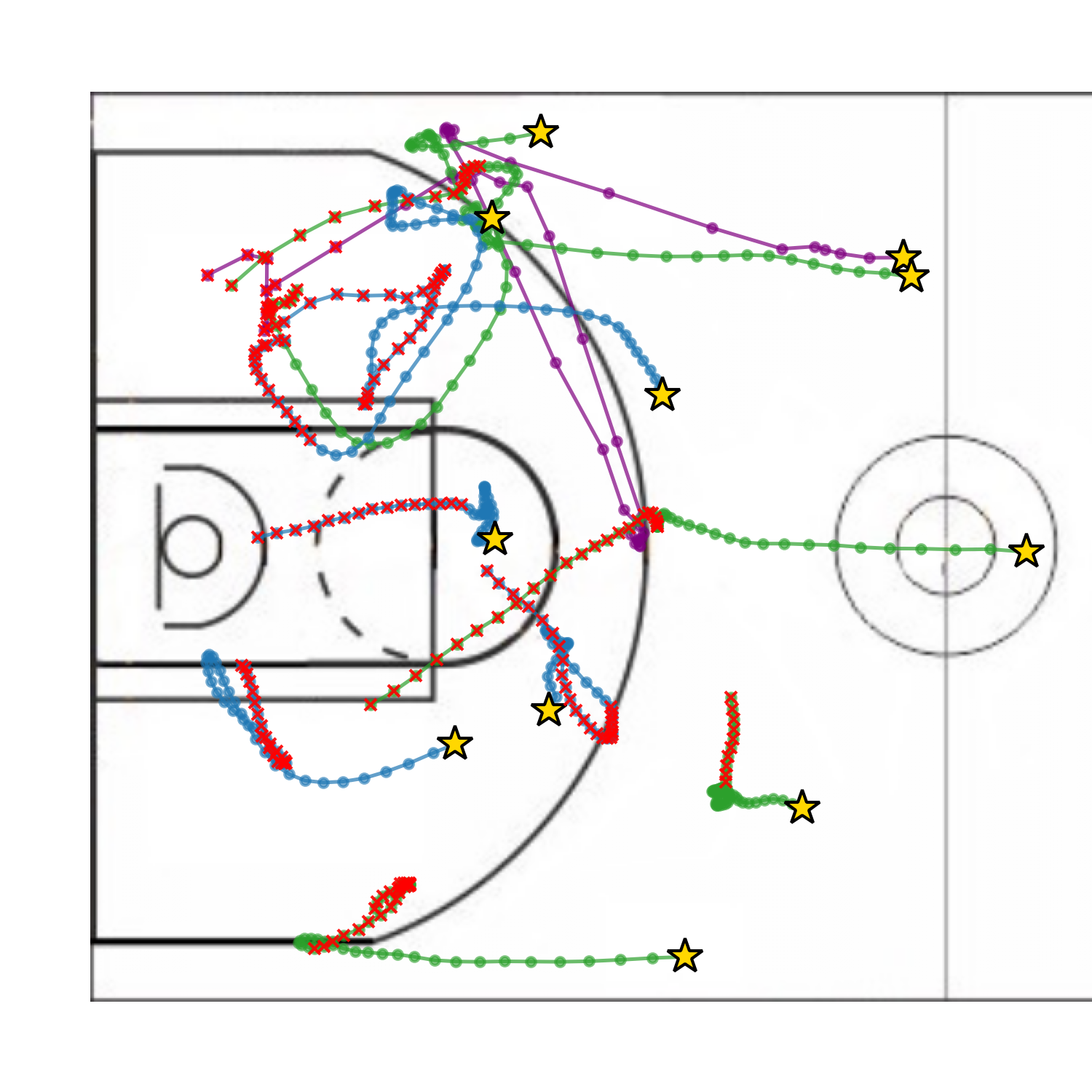}
        \caption{GT w/ Mask}
    \end{subfigure}
    \hfill
    \begin{subfigure}{0.32\linewidth}
        \includegraphics[width=\linewidth]{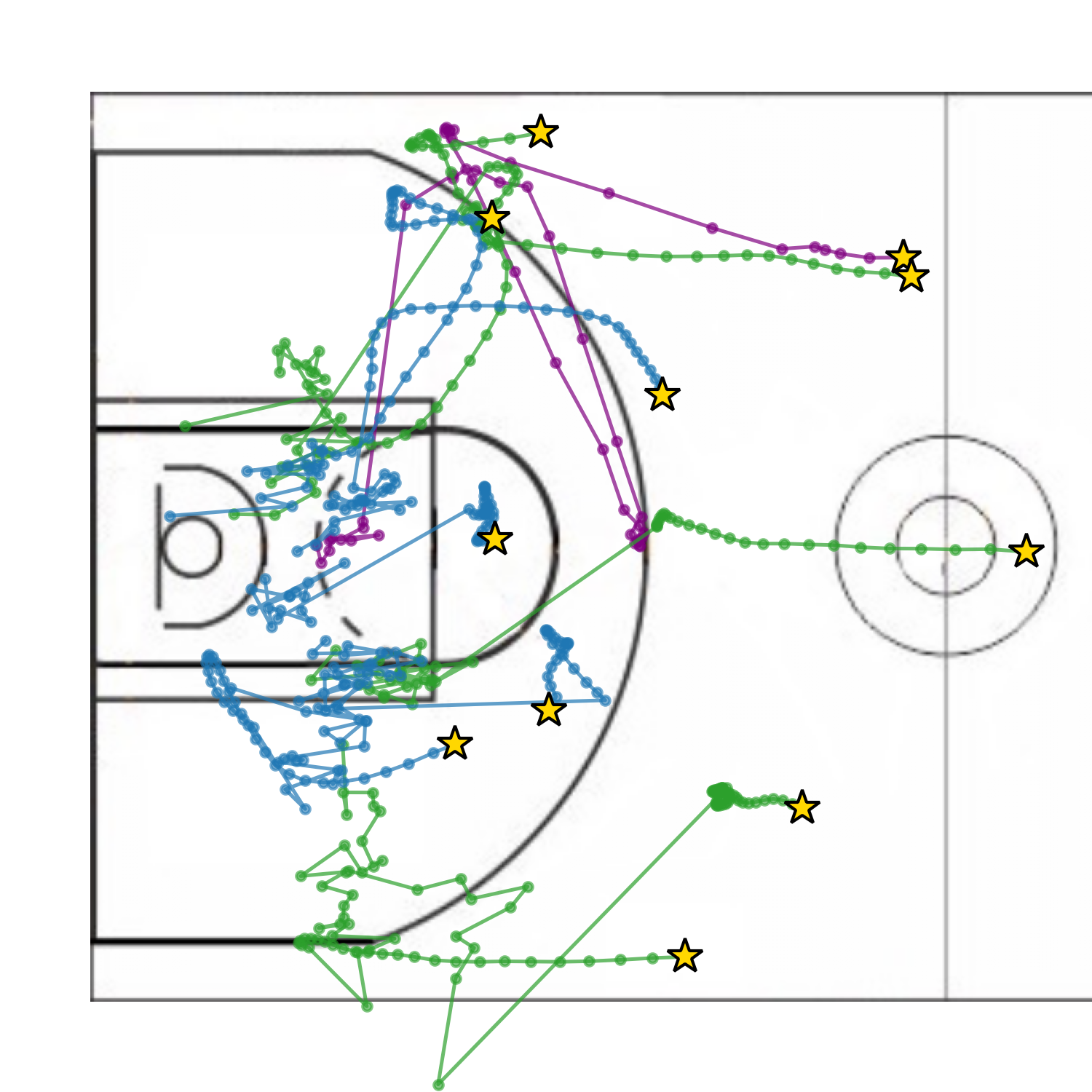}
        \caption{UniTraj}
        \label{fig:vis_e}
    \end{subfigure}
    \hfill
    \begin{subfigure}{0.32\linewidth}
        \includegraphics[width=\linewidth]{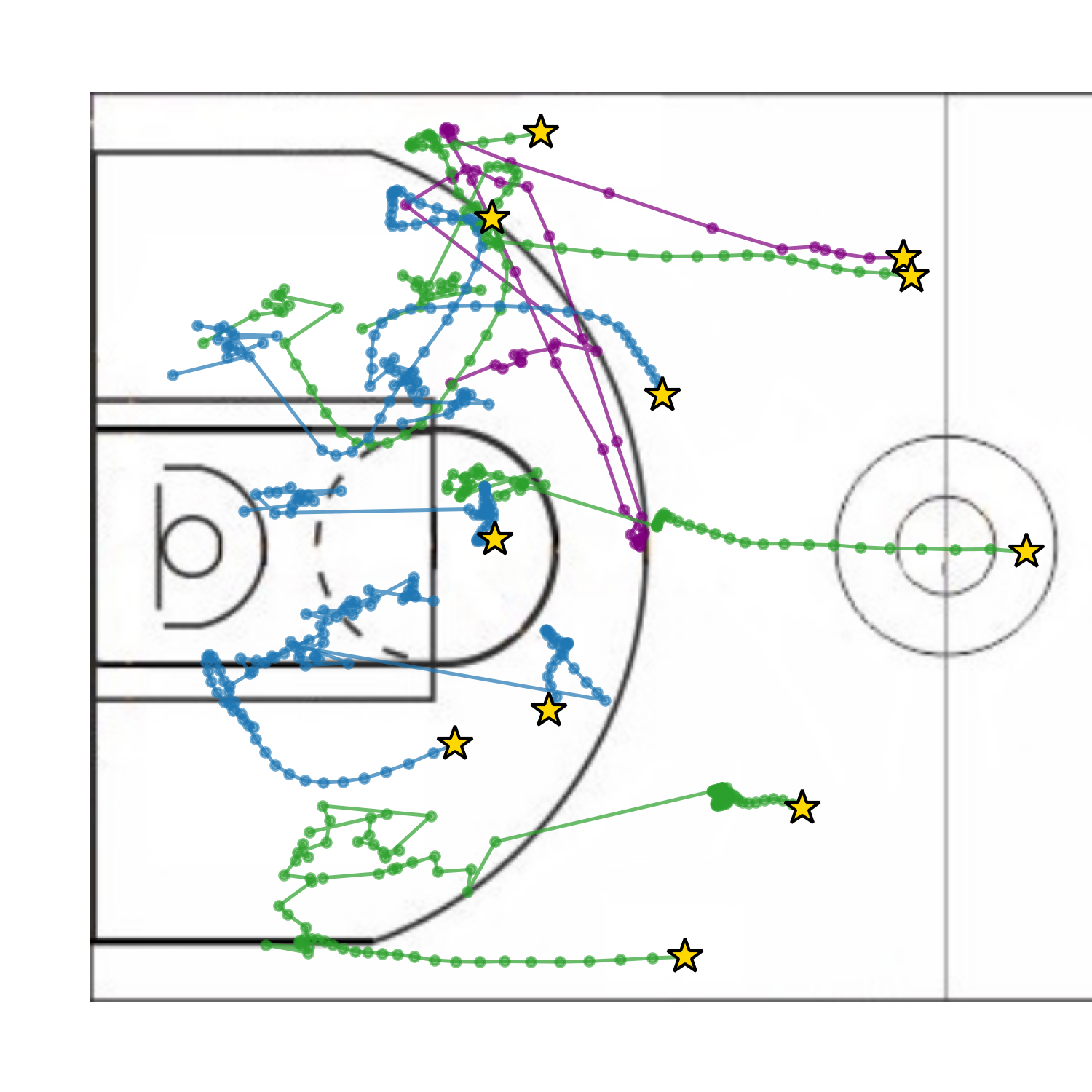}
        \caption{Ours}
    \end{subfigure}
    \caption{Qualitative comparison between UniTraj and our method on Basketball-U. Each row presents a different example, showing ground truth with mask (left), UniTraj prediction (middle), and our prediction (right).}
    \vspace{-3mm}
    \label{fig:U2S_basketball_vis}
\end{figure}

We further provide qualitative comparisons under the U2S setting in Fig.~\ref{fig:U2S_basketball_vis}. Compared to UniTraj, the trajectories produced by our model better approximate the ground truth and exhibit smoother motion patterns.
Notably, as shown in Fig.~\ref{fig:vis_e}, UniTraj occasionally generates trajectories that fall outside the basketball court boundaries or display unnatural zigzag paths. In contrast, our method produces more plausible movements, further validating the effectiveness of our approach.

%% file: table/S2S_basketball_football.tex
\begin{table*}[!t]
\centering
  \caption{Quantitative results on Basketball-U and Football-U datasets under the S2S setting (best highlighted, second-best underlined).}
  \label{tab:S2S_basketball_football}
  \scalebox{1.05}{
  \begin{tabular}{lccccc|ccccc}
    \toprule
    \multirow{2}{*}{Method} &  \multicolumn{5}{c|}{\textbf{Basketball-U} (In Feet)}  & \multicolumn{5}{c}{\textbf{Football-U} (In Yards)}\\
    \cmidrule(lr){2-11}
    & minADE$_{20}$ $\downarrow$ & OOB $\downarrow$ & Step & Path-L & Path-D 
    & minADE$_{20}$ $\downarrow$ & OOB $\downarrow$ & Step & Path-L & Path-D\\

    \midrule
  
    Mean 
    & 14.58 & \textbf{0} & 0.99 & 52.39 & 737.58 
    & 14.18 & \textbf{0} & 0.52 & 25.06 & 606.07\\
    Medium 
    & 14.56 & \textbf{0} & 0.98 & 51.80 & 743.36 
    & 14.23 & \textbf{0} & 0.52 & 24.96 & 600.22\\
    Linear Fit 
    & 13.54 & 4.47e-03 & 0.56 & 42.86 & 453.38
    & 12.66 & 1.49e-04 & \textbf{0.17} & \textbf{15.83} & 207.57 \\   
    LSTM~\cite{hochreiter1997long}
    & 7.10 & 9.02e-04 & 0.76 & 58.48 & 449.58
    & 7.20 & 2.24e-04 & 0.43 & 34.06 & 228.13 \\
    
    Transformer~\cite{vaswani2017attention}
    & 6.71 & 2.38e-03 & 0.79 & 59.34 & 517.54
    & 6.84 & 5.68e-04 & 0.42 & 33.01 & 202.10 \\
    
    MAT~\cite{zhan2018generating} 
    & 6.68 & 1.36e-03 & 0.88 & 58.83 & 483.46
    & 6.36 & 4.57e-04 & 0.40 & 31.32 & 186.11 \\
    
    Naomi~\cite{liu2019naomi} 
    & 6.52 & 2.02e-03 & 0.81 & 69.10 & 450.66 
    & 6.77 & 7.66e-04 & 0.67 & 42.74 & 259.11 \\ 
    
    INAM~\cite{qi2020imitative} 
    & 6.53 & 3.16e-03 & 0.70 & 58.54 & 439.87
    & 5.80 & 8.30e-04 & 0.39 & 32.10 & 177.04 \\
    
    SSSD~\cite{alcaraz2022diffusion} 
    & 6.18 & 1.82e-03 & 0.47 & 46.87 & 393.12
    & 5.08 & 6.81e-04 & 0.39 & 23.10 & 122.42\\
    
    GC-VRNN~\cite{xu2023uncovering} 
    & 5.81 & 9.28e-04 & 0.37 & 28.08 & 235.99
    & 4.95 & 7.12e-04 & 0.29 & 32.48 & 149.87\\

    UniTraj~\cite{xu2025sportstraj}
    & \underline{4.77} & 6.12e-04 & \underline{0.27} & \underline{34.25} & \underline{240.83}  
    & \underline{3.55} & 1.12e-04 & 0.23 & 19.26 & \underline{114.58} \\ 
    \cmidrule(lr){1-11}
    
    Ground Truth
    & 0 & 0 & 0.17 & 37.61 & 269.49
    & 0 & 0 & 0.03 & 12.56 & 76.68 \\
    \cmidrule(lr){1-11}
    \textbf{AdaSports-Traj}
    & \textbf{4.21} & \underline{4.25e-04} & \textbf{0.22} & \textbf{35.58} & \textbf{248.44}
    & \textbf{3.04} & \underline{0.88e-04} & \underline{0.20} & \underline{17.42} & \textbf{98.86}
    \\ 
    \bottomrule
  \end{tabular}
  }
  \vspace{-3mm}
\end{table*}

%% file: table/S2S_soccer.tex
\begin{table}[!t]
\centering
  \caption{Quantitative results on the Soccer-U dataset under the S2S setting (best highlighted, second-best underlined).
  }
  \label{tab:S2S_soccer} 
  \scalebox{0.88}{
    \begin{tabular}{lccccc}
    \toprule
    \multirow{2}{*}{Method} &  \multicolumn{5}{c}{\textbf{{Soccer-U}} (In Pixels)}  \\
    \cmidrule(lr){2-6}
    & minADE$_{20}$ $\downarrow$ & OOB $\downarrow$ & Step & Path-L & Path-D \\

    \midrule
    Mean & 417.68 & \textbf{0} & 4.32 & 213.05 & 8022.51\\
    Medium & 418.06 & \textbf{0} & 4.39 & 214.55 & 8041.98 \\
    Linear Fit & 398.34 & \textbf{0} & \textbf{0.70} & \textbf{112.34} & \textbf{2047.19}\\
    
    LSTM~\cite{hochreiter1997long}
    & 186.93 & 4.74e-05 & 7.50 & 652.98 & 4542.78\\
    
    Transformer~\cite{vaswani2017attention} 
    & 170.94 & 6.59e-05 & 6.66 & 566.14 & 4269.08 \\
    
    MAT~\cite{zhan2018generating} 
    & 170.46 & 7.56e-05 & 6.45 & 562.44 & 3953.34 \\
    
    Naomi~\cite{liu2019naomi} 
    & 145.20 & 8.78e-05 & 7.47 & 649.62 & 4414.99\\ 
    
    INAM~\cite{qi2020imitative} 
    & 134.86 & 4.04e-05 & 6.37 & 547.02 & 4102.37 \\
    
    SSSD~\cite{alcaraz2022diffusion} 
    & 118.71 & 4.51e-05 & 5.11 & 425.98 & 3252.66 \\
    
    GC-VRNN~\cite{xu2023uncovering} 
    & 105.87 & 1.29e-05 & 4.92 & 506.32 & 3463.26\\

    UniTraj~\cite{xu2025sportstraj}
    & 94.59
    & 3.31e-06
    & 4.52
    & 349.73
    & 2805.79 \\
    \cmidrule(lr){1-6}
    
    Ground Truth 
    & 0 & 0 & 0.52 & 112.92 & 951.00\\
    \cmidrule(lr){1-6}
    \textbf{AdaSports-Traj}  
    & \textbf{91.52}
    & \underline{2.14e-06}
    & \underline{3.87}
    & \underline{254.39}
    & \underline{2466.83}
    \\ 
    \bottomrule
  \end{tabular}
  }
  \vspace{-3mm}
\end{table}

%% file: table/U2S_basketball_football_soccer.tex
\begin{table}[!t]
\centering
  \caption{Performance comparison between UniTraj and our AdaSports-Traj when trained on Sports-U and evaluated on individual sports datasets. AdaSports-Traj achieves consistent improvements across all metrics and domains.}
  \label{tab:U2S_all} 
  \scalebox{0.88}{
    \begin{tabular}{lccccc}
    \toprule
    Method  & minADE$_{20}$ $\downarrow$ & OOB $\downarrow$ & Step & Path-L & Path-D \\
    \midrule
    & \multicolumn{5}{c}{\textbf{Sports-U to Basketball-U} (In Feet)} \\
    \cmidrule(lr){1-6}
    
    Ground Truth 
    & 0 & 0 & 0.17 & 37.61 & 269.49 \\
    UniTraj~\cite{xu2025sportstraj}
    & 11.12 & 3.82e-04 & 1.84 & 121.70 & 1684.26\\
    \textbf{AdaSports-Traj}
    & \textbf{8.74} & \textbf{2.76e-04} & \textbf{0.97} & \textbf{86.22} & \textbf{679.89}\\ 
     
    \bottomrule
    \toprule
    & \multicolumn{5}{c}{\textbf{Sports-U to Football-U} (In Yards)} \\
    \cmidrule(lr){1-6}
    
    Ground Truth
    & 0 & 0 & 0.03 & 12.56 & 76.68 \\
    UniTraj~\cite{xu2025sportstraj}
    & 11.89 & 1.92e-04 & 2.01 & 116.26 & 1549.79\\
    \textbf{AdaSports-Traj}
    & \textbf{9.04} & \textbf{1.24e-04} & \textbf{1.32} & \textbf{72.74} & \textbf{846.18}\\ 
    
    \bottomrule
    \toprule
    & \multicolumn{5}{c}{\textbf{Sports-U to Soccer-U} (In Pixels)} \\
    \cmidrule(lr){1-6}
    
    Ground Truth  
    & 0 & 0 & 0.52 & 112.92 & 951.00 \\
    UniTraj~\cite{xu2025sportstraj} 
    & 124.56 & 5.40e-05 & 13.64 & 677.72 & 9277.35\\
    \textbf{AdaSports-Traj}
    & \textbf{104.66} & \textbf{8.76e-06} & \textbf{6.78} & \textbf{486.91} & \textbf{6773.45}\\
    \bottomrule
  \end{tabular}
  }
\end{table}

%% file: table/ablation_module.tex
\begin{table}[!t]
\centering
\caption{Ablation study of AdaSports-Traj under both S2S and U2S settings. We investigate both the separate and joint impact of the Role- and Domain-Aware Adapter (RDA) and the Hierarchical Contrastive loss (HC). Results are reported in terms of minADE$_{20}$ on three datasets.}
  \scalebox{0.78}{
    \begin{tabular}{llccccc}
    \toprule
    \multirow{2}{*}{Setting} 
    & \multirow{2}{*}{Method}
    & \multirow{2}{*}{RDA} 
    & \multirow{2}{*}{HC} 
    &\multicolumn{3}{c}{minADE$_{20}$ $\downarrow$} \\
    \cmidrule(lr){5-7}
    &&&& \textbf{Basketball-U} & \textbf{Football-U} &  \textbf{Soccer-U} \\
      \midrule
    \multirow{3}{*}{\textbf{S2S}}
     & AdaSports-Traj & \checkmark &
     & 4.52 & 3.48 & 93.25\\
     & AdaSports-Traj & & \checkmark
     & 4.42 & 3.22 & 92.04\\
     & AdaSports-Traj & \checkmark & \checkmark 
     & \textbf{4.21} & \textbf{3.04} & \textbf{91.52}\\
     \bottomrule
    \toprule
     \multirow{3}{*}{\textbf{U2S}}
     & AdaSports-Traj & \checkmark &
     & 9.79 & 10.24 & 117.56\\
     & AdaSports-Traj & & \checkmark 
     & 9.34 & 9.81 & 112.85\\
     & AdaSports-Traj & \checkmark & \checkmark 
     & \textbf{8.74} & \textbf{9.04} & \textbf{104.66}\\
    \bottomrule
  \end{tabular}
  }
  \vspace{-3mm}
\label{tab:ablation_module} 
\end{table}

%% file: table/ablation_adapter.tex
\begin{table}[!t]
\centering
\caption{Ablation results on different adapter designs under both S2S and U2S settings. ``w/o gating'' uses only the attended latent $\boldsymbol{z}_{\text{cond}}$, ``w/ feature-wise'' applies a feature-wise gating without token-level modulation. Our ``w/ token-wise'' achieves the best performance across all domains.}
  \scalebox{0.85}{
    \begin{tabular}{llccc}
    \toprule
    \multirow{2}{*}{Setting} 
    & \multirow{2}{*}{Variant}
    &\multicolumn{3}{c}{minADE$_{20}$ $\downarrow$} \\
    \cmidrule(lr){3-5}
    && \textbf{Basketball-U} & \textbf{Football-U} &  \textbf{Soccer-U} \\
    \midrule
     \multirow{3}{*}{\textbf{S2S}} 
     & w/o gating 
     & 4.64 & 3.48 & 93.28\\
     & w/ feature-wise 
     & 4.68 &  3.50 & 94.12\\
     & \textbf{w/ token-wise (ours)} 
     & \textbf{4.52} & \textbf{3.48} & \textbf{94.25}\\
     \bottomrule
    \toprule
     \multirow{3}{*}{\textbf{U2S}} 
     & w/o gating 
    & 10.14 & 10.62 & 118.02 \\
     & w/ feature-wise 
     & 10.41 & 10.94 & 119.88\\
     & \textbf{w/ token-wise (ours)} 
     & \textbf{9.79} & \textbf{10.24} & \textbf{117.56} \\
    \bottomrule
  \end{tabular}
  }
\label{tab:ablation_adapter} 
\end{table}

%% file: table/ablation_contrastive_learning.tex
\begin{table}[!t]
\centering
\caption{Comparison of different contrastive learning strategies under the U2S setting. Our hierarchical contrastive learning outperforms others by separately supervising role- and domain-aware objectives using dedicated projection heads.}
  \scalebox{0.85}{
    \begin{tabular}{llccc}
    \toprule
    \multirow{2}{*}{Setting} 
    & \multirow{2}{*}{Variant}
    &\multicolumn{3}{c}{minADE$_{20}$ $\downarrow$} \\
    \cmidrule(lr){3-5}
    && \textbf{Basketball-U} & \textbf{Football-U} &  \textbf{Soccer-U} \\
    \midrule
     \multirow{4}{*}{\textbf{U2S}} 
     & Role-only
     & 10.02	& 10.51	& 114.68\\
     & Domain-only 
     & 10.39 & 10.87 & 114.29\\
     & Shared-feature
     & 14.86 & 12.15 & 136.54\\
     & \textbf{Hierarchical (ours)}
     & \textbf{9.34} & \textbf{9.81} & \textbf{112.85}\\
    \bottomrule
  \end{tabular}
  }
  \vspace{-3mm}
\label{tab:ablation_contrastive} 
\end{table}

%% file: section/6_conclusion.tex
\section{Conclusion}
We propose AdaSports-Traj, a unified and adaptive framework for multi-agent trajectory modeling in sports. 
In addition to strong performance under the standard single-domain (S2S) setting, we propose a new cross-domain evaluation protocol (U2S) to better assess generalization across different sports. Our Role- and Domain-Aware Adapter and Hierarchical Contrastive Learning modules enable the model to explicitly capture structural differences across agent roles and sports domains, effectively mitigating distribution shifts and improving generalization in both in-domain and cross-domain settings.

\noindent\textbf{Limitations and Future Work.}
While AdaSports-Traj demonstrates strong generalization across roles and domains, it currently relies on pre-defined role and domain labels during training. In future work, we plan to explore label-free or weakly supervised variants to enhance scalability. 